\documentclass[10pt, a4paper]{article}

\usepackage[final]{lrec2026} 

\usepackage{csquotes}
\usepackage{booktabs}
\usepackage{amsmath}
\usepackage{multirow}
\usepackage{float}  
\usepackage{array}
\usepackage{nicefrac}
\newcommand{\rowstrut}{\rule{0pt}{2.6ex}} 

\usepackage{fontspec}

\newfontfamily{\junicodeFont}{Junicode-Regular.ttf}

\newcommand{\character}[1]{{‹{\junicodeFont{}#1}›}}

\newif\iffinal
\finaltrue       
\newcommand{\finalonly}[1]{\iffinal #1\fi}
\title{Pre-Editorial Normalization for Automatically Transcribed Medieval Manuscripts in Old French and Latin}

\name{Thibault Clérice,$^1$ Rachel Bawden,$^1$ Anthony Glaise,$^2$ \\\textbf{\large Ariane Pinche},$^3$ \textbf{\large David Smith}$^4$} 

\address{
         $^1$Inria, Paris, France $^2$CNRS (UMR 7323), Tours, France, \\ $^3$CNRS (UMR 5648), Lyon, France, $^4$Northeastern University\\
         \texttt{\{firstname.lastname\}@\{inria,cnrs\}.fr}, \texttt{dasmith@ccs.neu.edu}
         }

\abstract{
Recent advances in Automatic Text Recognition (ATR) have improved access to historical archives, yet a methodological divide persists between palaeographic transcriptions and normalized digital editions. While ATR models trained on more palaeographically-oriented datasets such as CATMuS have shown greater generalizability, their raw outputs remain poorly compatible with most readers and downstream NLP tools, thus creating a usability gap. On the other hand, ATR models trained to produce normalized outputs have been shown to struggle to adapt to new domains and tend to over-normalize and hallucinate. We introduce the task of Pre-Editorial Normalization (PEN), which consists in normalizing graphemic ATR output according to editorial conventions, which has the advantage of keeping an intermediate step with palaeographic fidelity while providing a normalized version for practical usability. We present a new dataset derived from the CoMMA corpus and aligned with digitized Old French and Latin editions using \textit{passim}. We also produce a manually corrected gold-standard evaluation set. We benchmark this resource using ByT5-based sequence-to-sequence models on normalization and pre-annotation tasks. Our contributions include the formal definition of PEN, a 4.66M-sample silver training corpus, a 1.8k-sample gold evaluation set, and a normalization model achieving a 6.7\% CER, substantially outperforming previous models for this task.
 \\ \newline \Keywords{Latin, Old French, Normalization, Text Recognition, Medieval manuscripts} }

\begin{document}

\maketitleabstract

\section{Introduction}

\begin{figure}[h]
    \centering    
    \resizebox{\linewidth}{!}{%
    \begin{tabular}{l|c}
        \multicolumn{2}{c}{\includegraphics[width=\linewidth]{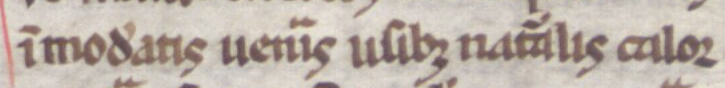} }\\[0.5em]
        Graphetic & {\large \junicodeFont ı̄mo̕atis uenı̈s uſibꝫ nt᷑lıs cloꝛ} \rowstrut\\
        Graphemic & {\large \junicodeFont ĩmod̾atis uenĩs usib nat᷑alis calor} \rowstrut\\
        Normalized & {inmoderatis veneris usibus naturalis calor} \rowstrut\\
    \end{tabular}}
    \caption{The outputs of different transcription approaches for the same line of text. Manuscript: BSB CLM 13027. The task defined in this paper consists in producing normalized transcriptions from graphemic outputs.}
    \label{fig:transcription_type}
\end{figure}

The progress of Automatic Text Recognition (ATR) has facilitated access to historical archives \cite{jdmdh:13247}. However, there exists a methodological divide in how to transcribe manuscripts on a spectrum ranging between two extremes: the graphetic approach and a normalized approach (Figure~\ref{fig:transcription_type}). The graphetic approach, defined by \citet{robinson1993guidelines}, treats transcription as a graphical representation of the manuscript's text, which captures each character variant, and maintains original punctuation and abbreviation without expansion. Normalized transcription conflates transcribing with textual editing~\cite{pierazzo_rationale_2011}, thus adhering more to users expectations by \enquote{[...] 
[making] the manuscript text accessible to the modern reader by expanding abbreviations, separating words, inserting diacritical marks, and adding modern punctuation}~\cite[p.330]{duval}. However, this approach, combining two tasks at the same time, a graphical interpretation and a linguistic  interpretation, has shown to poorly resolve abbreviations and to hallucinate content~\cite{aguilar2024handwritten,torresaguilar:hal-03892163}, without providing an ability for the reader to distinguish normalization errors from graphical interpretation errors.

It can be useful to follow an approach between the two extremes, whereby character variants are merged but where abbreviations and other typical manuscripts phenomena are preserved: this is known as the graphemic approach. This enables the transcriptions to be treated as their own epistemic objects, which is an advantage in itself. It also results in transcriptions that result in models, such as those trained on the CATMuS dataset~\cite{catmus}, that better generalize to new domains and languages~\cite{coll2025evaluating}. 

However, graphemic transcriptions create a ``usability gap'' where the most accurate transcriptions of the material reality of the text are the least compatible with most available tools for computational analysis and expectations from the humanities~\cite{kiessling:hal-05429033}.\footnote{Tools such as lemmatizers, part-of-speech (PoS) taggers, and Named Entity Recognition (NER) systems are typically trained on normalized, modern-style editions. When faced with the raw, non-resolved output of a graphemic ATR model, the performance of these tools drops significantly \cite{wauchieristhatyou}.} It can therefore be useful to automatically produce normalized versions of these transcriptions. It holds an advantage over all-in-one normalized ATR approaches, as it provides an intermediate layer that makes it possible to distinguish ATR error (such as \character{rn}/\character{m}) from normalization errors (\character{ẽ\&} for \character{est et} or \character{esset}).

In this paper, we introduce the task of automatically producing normalized transcriptions from graphemic ATR transcriptions, which we name pre-editorial normalization (PEN). The task definition and rationale is followed by the introduction of a novel dataset, based on the ATR silver corpus CoMMA~\cite{comma} and digitized and born-digital editions of Old French and Latin texts. Using \textit{passim}~\cite{10.1093/alh/ajv029}, we automatically aligned both corpora, of which we then manually corrected some of the alignments to produce a gold-standard test set. We benchmark the silver dataset using models trained on a sequence-to-sequence transformer, ByT5 \citep{xue-etal-2022-byt5}, across two tasks: (i)~a normalization task, for which we consider our gold test set as the goal, and (ii)~a pre-annotation task for lemmatization and PoS-tagging.


\section{Related Work}

\subsection{Automatic Text Recognition with Normalized Output}

ATR research has generally paid limited attention to the distinction between imitative transcription of historical sources and their subsequent normalization. Historical text recognition models are often evaluated on resources such as the IAMHistDB medieval datasets, the {Perzival} \cite{partzival2009fischer}, and the {Saint Gall Database} \cite{stgall2011fischer}, despite the fact that these corpora rely on different transcription paradigms. \citet{camps:hal-03916914} benchmarked both approaches on genre-constrained documents but without providing detailed qualitative analyses of the results.

By contrast, several studies within the Digital Humanities community have explicitly addressed the risks associated with over-normalization. In particular, two recent works have examined this issue as a central research question. \citet{bottaioli2024normalized} show that models trained to jointly transcribe and normalize middle names in digitized birth certificates may hallucinate names for individuals who did not possess one. Similarly, in the context of medieval manuscripts and, more broadly, in languages with highly variable orthography, \citet{torteroloorta2025transcribingspanishtextspast} demonstrate that increased accessibility through normalization entails a latent risk of systematically erasing historically and linguistically meaningful variation.

Conversely, qualitative error analyses conducted in several studies by S.~T. Aguilar highlight persistent limitations of models trained on normalized outputs, particularly with respect to abbreviation resolution. \citet{torresaguilar:hal-03892163} show that morphosyntactic errors constitute one of the most frequent sources of abbreviation-related errors. Moreover, \citet{aguilar2024handwritten} report several critical failures, including difficulties in handling line-end hyphenation, the misinterpretation of simple abbreviations (e.g.,~\character{ĩfusũ} for \character{infusum} transcribed as \character{in futurum}) even in relatively straightforward contexts, and errors in local semantic disambiguation, such as interpreting \character{diaꝰ} as \character{diaconus} `deacon' instead of \character{diabolus} `devil'.

\subsection{Historical Text Normalization}

Datasets for historical text normalization remain scarce, particularly for Latin. To our knowledge, the only comprehensive and readily available non-synthetic dataset for Latin historical normalization is the work of \citet{schonhardt_bdd_abbreviations_2025}, which is limited to a single text attested in six manuscripts. This restricted scope limits its suitability for studying abbreviation expansion in a broader diachronic and bilingual context. A small number of additional datasets have been produced or identified in the work of \citet{camps:hal-03916914}, some of which include Old French. However, their coverage remains limited in terms of genre and textual diversity.

Related work by \citet{buttner_byt5-base-latin-normalize_2025} addresses the normalization of Latin spelling by aligning medieval forms with more classical orthographic standards using a ByT5-based approach. While this study provides a relevant example of normalization in the Latin domain, it focuses on orthographic regularization rather than pre-editorial normalization.

By contrast, normalization for vernacular languages has received greater attention, beginning with late Middle French and extending into the early modern period. \citet{solfrini-etal-2025-normaliser} present a study on the normalization of printed late Middle French texts from the 1530s. More broadly, the automatic standardization of late Middle French has been explored in several works \cite{rubino-etal-2024-normalizing,rubino-etal-2024-automatic}. These approaches are similar in methodology but differ in purpose from ours, as their primary aim is to reduce endemic linguistic variation rather than to preserve philological and historical features. On this topic, \citet{bawden-etal-2022-automatic} compare normalization models for early modern French (17th c. and later), including datasets and comparative assessments of statistical and neural machine translation approaches.

Finally, digital editions may encode such information using a variety of representation frameworks, as illustrated in doctoral work \cite{franzini2019towards} and other projects \cite{nury_2024}. However, beyond the technical challenges involved in interpreting and exporting data from these platforms, identifying suitable editions remains a difficult task. 
To date, the only available catalogue providing filters for downloadable TEI resources \cite{andorfer_franzini_2025} indicates that, among more than 29 digital editions with accessible XML-TEI files for Latin and French texts, none preserves abbreviated forms with explicit markup.\footnote{\citet{DigitalScholarlyEditions2020} does not provide such metadata.}

\subsection{Normalization of User-Generated Content (UGC)}


Another type of data associated with the lexical normalization task is user-generated content (UGC), of which social media texts are the most prevalent \cite{Baldwin2013-dm}. UGC, like the historical data dealt with in this article, is characterised by non-standard phenomena such as acronyms, abbreviations, truncations and variations in spelling. Such texts can present challenges for natural language processing (NLP) tools, particularly those trained on standard, edited texts, which has led to a body of work in normalisation as a preprocessing step \cite{melero-etal-2012-holaaa,baldwin-etal-2015-shared,van-der-goot-etal-2021-multilexnorm,nishimwe-etal-2024-etude}.

Initial approaches relied on rules, lexicons and language models, and it was common to decompose the task into the detection of words to be modified and their normalisation \cite{baranes-normalisation-2015,ikeda-etal-2016-japanese}.
Later approaches opt for a translation-like approach, where the task is treated as the translation of a non-standard to standard language sequence \cite{aw-etal-2006-phrase,declercq-normalization-2013}. This approach has the advantage of increased use of context with the sequence and flexibility, helpful notably for segmentation and reordering differences.

The choice of representation has been a subject of much research in the handling of UGC texts (both for normalisation and other NLP tasks). Phonetic representations have been explored \cite{rosales-nunez-etal-2019-phonetic,mandal-nanmaran-2018-normalization}, particularly for spelling errors. Character representations have also shown promise in terms of increasing robustness against phenomena such as spelling errors, devowelling, acronyms and abbreviations \cite{belinkov2018synthetic,watson-etal-2018-utilizing,domingo-comparison-2021,scherrer-ljubesic-2021-sesame}. Character-level representations have the advantage of better capturing similarities between similar but non-identical lexical forms. It is an approach that has previously been shown to work well for morphologically rich languages (which are also characterised by a higher level of variation) and low-resource scenarios (in order to generalize better across forms). For this reason, another successful approach is to either collect and annotate more data or to generate synthetic data \cite{belinkov2018synthetic,vaibhav-etal-2019-improving,nishimwe-etal-2024-making}. 

Recent work has shown that models trained on large quantities of data, and in particular large language models (LLMs) are already ``more robust'' in that they perform better than previously on non-standard UGC data, without being specifically trained to do so and without having to apply a specific normalization pre-processing step \cite{bawden-sagot-2023-rocs,peters-martins-2025-translation,bawden-sagot-2025-rocs}.\footnote{Given that they are trained on large amounts of web data, it can be debated whether the models are actually more robust or whether the non-standard forms are simply sufficiently common in the training data.} However, this is unlikely to be the case across all languages, and in particular low-resource ones.


\subsection{Automatic Alignment}

Aligning multiple data representations has been an important method for dataset creation since the application of dynamic time warping in speech recognition \cite{Sakoe1978DynamicPA} or dynamic programming for aligning ATR outputs or multiple translations of the same source \cite{pial_gnat_2023,thai_exploring_2022}.  This paper uses methods developed for text-reuse detection, which is a broad task with many possible approaches and evaluations \cite{moritz_reuse_2016}.  As described below (\S\ref{sec:alignment}), the main inference task in creating our dataset is to match and align the ATR output of manuscript pages with passages in normalized editions.  Unlike in other ATR evaluation tasks, these normalized editions do not have page breaks that match any particular manuscripts and the manuscripts are often miscellanies with several works.  The size of the corpus makes it impractical to compare all pairs of ATR pages and passages from digital editions \cite{paju_ontology_2022}.  We therefore employ passim 2.0\cite{smith23:chr}, which uses hashed character $n$-gram fingerprinting to filter the set of candidate passages for alignment. Once a pair of passages is selected, passim uses a character-level hidden Markov alignment model \cite{vogel_hmm-based_1996}.  Unlike text-reuse packages that use the Smith-Waterman alignment algorithm \cite{smith14:jcdl,funk_spine_2018}, the HMM approach is more robust to the rearrangement of passages in the ATR output due to uncertainty in page layout analysis.

\section{Pre-editorial Normalization}

\subsection{Task Definition}\label{sec:taskDef}

The output of graphemic ATR\footnote{\citet{robinson1993guidelines} defines graphemic as a transcription keeping original practices such as abbreviation but merging variants of same characters, such as \character{s} and \character{ſ}. They are implemented and more broadly defined in \cite{catmus}.} is inherently raw and reflects both manuscript-specific phenomena and transcription guidelines. As a result, it exhibits several sources of variability: spacing is inconsistent (including but not limited to unmarked line-break hyphenation), capitalization---especially in earlier centuries---may serve a decorative rather than linguistic function, numerical expressions vary across systems (ranging from Roman numerals to fractional notation in later periods), ramist letter pairs (\character{u/v} and \character{i/j}) are not systematically distinguished, punctuation practices frequently differ from those adopted in modern editorial standards, and abbreviations are preserved. Finally, especially in the context of raw ATR output, recognition errors are superimposed on top of the multiple sources of variation described above.

These characteristics considerably limit the direct usability of ATR outputs in standard NLP and text-processing pipelines. Moreover, they diverge substantially from the normalized representations typically expected by scholars in the humanities, thereby motivating the need for dedicated normalization and post-processing methods.

The PEN task consists in:
\begin{itemize}
    \item normalizing letters that were considered allographs in the middle ages and that diverged over time, such as ramist letters (\character{u}/\character{v}) in consonantal and vocalic usage. Historical spelling variants are not normalized; for instance, \character{f} replacing an etymological \character{ph} is preserved.
    \item resolving abbreviations in a context-sensitive manner: expansions should reflect not only the semantic content of words, but also their morphosyntactic features and a coherent orthographic system within the text.
    \item addressing recognition errors in the context of raw ATR output, such as \character{rn}/\character{m} confusions.
    \item normalizing capitalization around punctuation and proper names according to modern editorial standards.
    \item preserving original punctuation.
\end{itemize}

\subsection{Limits of the Task}

Certain abbreviations, especially those referring to personal names or place names, should not be expanded when the available knowledge in context is insufficient to support a reliable interpretation. 
For example, the abbreviation \character{G.\ dixit} `G. said' in a medical text can be reliably expanded to \character{Galenus dixit} based on contextual evidence. On the other hand, in a charter or accounts, there might not be any contextual elements to resolve the name of S. in \character{S. Parisiensis} `S. the Parisian', which could be many things (Stephanus, Severinus, etc.) and the initial should therefore be kept as it is.

Severe or unprocessable ATR errors resulting in extended spans of noise should not be corrected by the model. Attempting to normalize such segments risks introducing over-normalization and, more critically, hallucinated content. Given the text-based nature of the task, only recoverable textual information should be considered.

Numerical expressions written in Roman or Arabic numerals are likewise left unchanged. This form of variation remains an open issue within the community and requires further standardization before being fully integrated into the task definition. Moreover, the distinction between ordinal and cardinal numbers is often expressed through contextual or morphological markers and may require complex morphosyntactic analysis to be resolved reliably on top of the inherent variation of Roman numeral spelling.

Sufficient contextual information is required at all levels (semantic, morphosyntactic, and spelling) to support reliable normalization decisions. However, contextual evidence may itself be conflicting. For instance, if both \character{molt} and \character{moult} occur within the same Old French text, it remains unclear how the abbreviated form \character{młt} should be expanded. 

In this paper, we frame the task as a single input--single output problem. In future work, the PEN task should also involve a processing layer capable of representing a distribution over possible outputs, to account for spelling variations within a single document, which can be seen in Old French (\character{segnor} and \character{signor} on the same page).

\section{Dataset Construction}

\subsection{Source Data}

Our data for historical transcriptions is drawn from the CoMMA corpus, a graphemic-transcription corpus of 33,000 manuscripts (3.3B whitespace-delimited tokens). In our approach, we use the main text of each page to build full-text manuscript pages, ignoring running headers, marginal zones, and interlinear comments, thanks to the dataset's fine-grained details concerning layout classification. While manuscripts come with language metadata, we do not filter based on it, as it represents document-scale language the metadata: our alignment could find excerpts of Latin or Old French in manuscripts not annotated as such.

For the normalized texts to align with CoMMA, we use publicly available data for Latin, and a mix of closed and open data for Old French, to ensure a large enough pool of data to align with the manuscripts. Both the Latin pool and the Old French one are drawn from the same data cited by CoMMA for their ModernBert model \cite[Table~6]{comma}. Some of the corpus contains highly technical data (mathematics, medical, etc.), which we expect to be the hardest to normalize in Latin, as it contain higher levels of special vocabulary and specific abbreviations.

\subsection{Alignment}
\label{sec:alignment}

We align the ATR transcripts of the CoMMA corpus with the plain text of digital editions of Latin and Old French texts using passim 2.0 \cite{smith23:chr}.  This version of the package implements a character-level HMM alignment algorithm \cite{vogel_hmm-based_1996}, which permits rearrangement of passages between two versions of a text.  This is helpful in working with ATR output, when the sequence of page regions in the transcript does not match the sequence as transcribed in a scholarly edition. In contrast to other ATR ground-truth datasets produced by forced alignment, we do not need to specify in the input which manuscripts are versions of which texts; rather, we specify to passim that each page from CoMMA should be aligned to one or more pages from the corpus of digital editions.  We performed unicode decomposition normalization on both the CoMMA ATR output and the digitial editions so that the common use of diacritics for, e.g., nasalization (\character{cõsul} for \character{consul}) or truncation (\character{\'{c}} for \character{cum}) would align more closely to their standard forms. To improve recall, we index the collection with character 10-grams (ignoring whitespace and punctuation) that occur in fewer than 100 pages. It then aligns pages having at least five 10-grams in common and drops alignments shorter than 50 characters.  We use beam search ($b = 200$) when aligning pages.

From the automatic alignment, we filter the data with at least five lines of continuous aligned content, a sufficiently large match rate (60\%), and only where the alignment comes from the same source text. This resulted in 3.2M alignments. We then reprocessed them into smaller alignments, with a maximum and minimum size of 1000 and 300 bytes respectively, to get down to a total of 4,665,509 pairs.

\subsection{Limits}

As our approach relies on the automatic alignment of raw ATR output with digitized editions, several sources of noise may affect the resulting annotations regarding the task definition. First, since the distance between an abbreviated token and its expanded form influences alignment quality, there is a non-negligible probability that a manuscript form is aligned with a different lexical variant in the critical edition. This situation is common in textual transmission and may result in semantically or syntactically divergent alignments. Additionally, over-normalization may also occur, in a manner similar to variant-swapping phenomena, such as \character{f}/\character{ph} alternations in Latin or \character{moult}/\character{molt} variation in Old French. These cases may lead to normalized forms that contradict the intended task definition.

\subsection{Description}

Unsurprisingly, classical, late-antique, and Christian ``literary'' Latin texts are over-represented in the alignments between our source corpora and CoMMA, while later medieval administrative texts are under-represented. This reflects the fact that CoMMA is largely derived from library collections rather than archives. Patrologia Latina, our largest source (97M tokens, \citetlanguageresource{CC}), accounts for 64.5\% of the alignment pairs. By contrast, CEMA, an archival corpus and the second-largest source (57M tokens, \citetlanguageresource{perreauCEMA}), represents only 1.5\% of the alignments, whereas Perseus (6.1M tokens, \citetlanguageresource{perseus}) accounts for 6.5\%.

As expected from studies on manuscript transmission, some works are much more represented than others (see Table~\ref{tab:most_freq_work}), and all most common authors are either tied to religious practice (missals, sermons and homilies), theory (commentaries),  or canon law (\textit{decretum}). One reason for this over-representation of some texts might be the presence of numerous bible citations.

\begin{table}[]
\resizebox{\linewidth}{!}{
\begin{tabular}{@{}llr@{}}
\toprule
Corpus & Work                                                 & \# Mss. \\ \midrule
PL     & Auctor Incertus - Missale mixtum                     & 2090        \\
PL     & Gratianus - Concordia discordantium canonum          & 2026        \\
encyMe & Speculum doctrinale                                  & 2002        \\
CSEL   & Augustine - Liber qui appellatur Speculum            & 1795        \\
PL     & Martinus Legionensis - Sermones          & 1735        \\
PL     & Auctor incertus - Breviarium             & 1625        \\
PL     & Ivo Carnotensis - Decretum               & 1613        \\
\bottomrule
\end{tabular}}
\label{tab:most_freq_work}
\caption{Top works by manuscript presence.}
\end{table}

Over-normalization, in the form of variant insertion, spelling standardization (in Latin), or lexical replacement (in Old French), constitutes one of the main identified risks. At this stage, however, we are unable to reliably distinguish over-normalization from two related phenomena: first, cases in which abbreviation markers are not recognized by the ATR system (e.g. \character{ifusa} for \character{ĩfusã} with missed diacritics), and second, post-ATR editorial corrections, which are likely more prevalent.

\begin{figure}
    \centering
    \includegraphics[width=\linewidth]{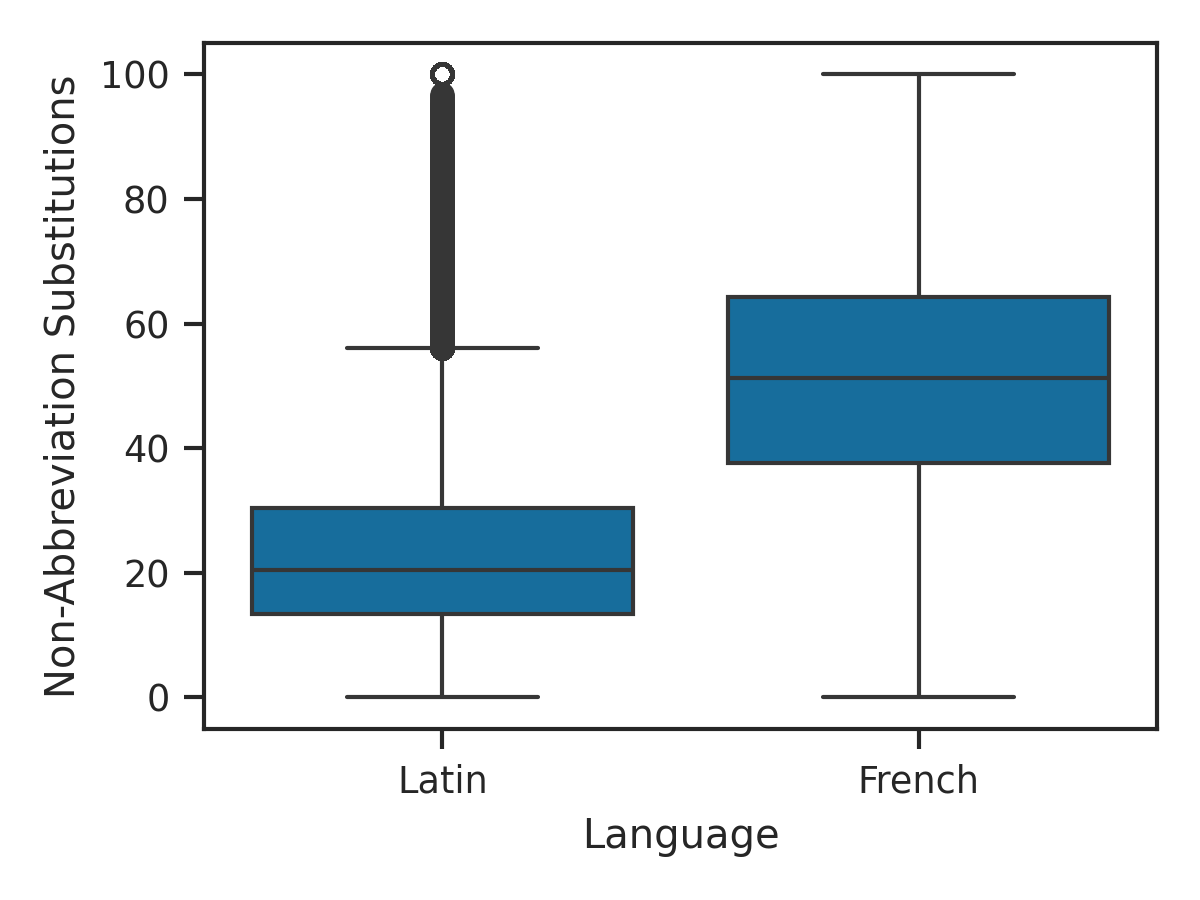}
    \caption{Distribution of the percentage of tokens per alignment in the train set that do not correspond to abbreviation resolution pattern.}
    \label{fig:train_insertion}
\end{figure}

Nevertheless, we estimate the extent of over-normalization by automatically aligning tokens in the training set and measuring the proportion of tokens that differ between the source and aligned texts when no explicit abbreviation markers are present. This analysis reveals a marked contrast between the two languages (Figure~\ref{fig:train_insertion}). In Latin, 20.5\% of tokens exhibit substitutions consistent with non-standard abbreviation resolution. This proportion broadly aligns with expectations based on the 9.7\% character error rate (CER) reported for CoMMA, including additional effects from the limited possibilities of over-normalization in Latin and variant substitutions. 

By contrast, Old French displays a substantially higher level of non-abbreviation related substitutions, with a median substitution rate of 51\% (and a mean of 50\%). This discrepancy suggests that, compared to Latin, Old French editions impose a greater degree of spelling normalization on manuscript sources, thereby introducing more extensive spelling variation between manuscripts and their corresponding editorial representations.

\subsection{Manual Correction and Gold Standard}

To create a gold set, we sampled some data to be representative of main corpora from the automatic alignment. The resulting pairs are all manually corrected by a philologist to resolve alignment errors, normalization inconsistencies, and segmentation issues. This  results in a gold-standard dataset that serves both for evaluation and for qualitative error analysis. To that test data, we add ground truth data from CATMuS data repositories (both archives and literary documents) that was manually normalised by the same expert.

In the automatically aligned ATR output-part of the gold dataset, two texts are substantially overrepresented: \textit{Oresme} (75 alignments) and the \textit{Histoire Ancienne} (43). This imbalance reflects the limited diversity of available Old French datasets, which made it challenging to construct a fully balanced sample. All other corpora are represented by between 11 and 21 alignments. Among these, some technically demanding texts are included, such as the \textit{Mathematica} corpus. It is represented by 20 alignments and contains atypical elements such as geometrical point labels (e.g. `A'), which resemble abbreviations but must not be treated as such.

The ground truth ATR focused specifically on two series of texts: the first are drawn from archives, small letters and charters, mostly from the Cartularies of Notre-Dame de Paris~\cite{claustre2022ndp}; the second are drawn from a single manuscript bearing the text \textit{Liber de Coitu} (Constantine the African, \textsc{xi} c.). The first series contains administrative text, with various anonymous people and occasional switching between Latin and French, while the second contains very abbreviated content, such as \enquote{\junicodeFont eliqͣt᷑ hũor qͥ ẽ.ĩ ce̾b ͦ ⁊ eliqͣtꝰ} (corresponding to \textit{eliquatur humor qui est in cerebro, et eliquatus} `the humor that is in the brain liquefies, and once it is liquefied').

The final gold dataset comprises 588 samples (64,230 tokens) from 316 different manuscripts. 47 alignments were drawn from ground truth ATR, with no or very few recognition errors, and which are therefore longer than other samples (287 tokens on average vs.~94, see Table~\ref{tab:distrib-gt}). 

\begin{table}[]
\resizebox{\linewidth}{!}{
\begin{tabular}{@{}llrrrr@{}}
\toprule
                 &        &                      & \multicolumn{3}{l}{\textbf{Word Length}}          \\ \midrule
                 &        & \textbf{Samples}     & \textbf{Average} & \textbf{Std} & \textbf{Median} \\ \midrule
\textbf{Raw ATR} & French & 189                  & 95               & 34           & 89              \\
                 & Latin  & 352                  & 93               & 39           & 88              \\ \midrule
\textbf{GT ATR}  & Latin  & 45                   & 288              & 256          & 259             \\
                 & Mixed  & 2                    & 276              & 101          & 276             \\ \bottomrule
\end{tabular}}
\label{tab:distrib-gt}
\caption{Distribution of languages, sample counts, and token counts across the two alignment sources: raw ATR output and ground truth (GT).}
\end{table}

\section{Benchmarking Normalization}

Our first experiment evaluates the impact of the proposed dataset in the context of ATR output normalization. Specifically, we train a model for abbreviation resolution and compare its performance against off-the-shelf approaches, including zero-shot and domain-limited pretrained models. 

\subsection{Setup}

We formulate text normalization as a sequence-to-sequence task. We select ByT5 as the base architecture, as it can handle a wide range of characters (including Unicode-decomposed forms), since it operates at the byte level. This property has been associated with improved performance on languages exhibiting high levels of spelling variation \cite{wolters2024historical}. ByT5 has also been successfully applied to related tasks such as ATR correction \cite{pavlopoulos-etal-2024-challenging}, medieval Latin normalization \cite{buttner_byt5-base-latin-normalize_2025}, and abbreviation expansion \cite{schonhardt_bdd_abbreviations_2025}. 

\paragraph{Baselines}
We consider two baseline systems. The first is a ByT5-based model trained on the dataset of \citet{schonhardt_bdd_abbreviations_2025}. This model relies on different Unicode-level guidelines for abbreviations and is strongly domain-constrained. Despite the shared scribal conventions for abbreviation between Old French and Latin, we expect this model to perform poorly on Old French, which is out of domain. In addition, the dataset's source text were manually corrected, which may hinder robustness to ATR noise and to inconsistencies in spacing. 

Our second baseline consists of zero-shot prompting Llama~3~8B Instruct \cite{grattafiori2024llama3herdmodels}. The model is prompted to perform PEN, with an auxiliary instruction to predict the input language, and common abbreviation patterns and glyphs\finalonly{ (Appendix~\ref{app:prompt})}.

\paragraph{Training data}
The training data is bilingual, combining Latin and Old French. Owing to a strong imbalance between the two languages and the relative scarcity of Old French data, we upsample the French portion of the training set by a factor of ten, reducing the imbalance from approximately 100:1 to 10:1 (Latin to Old French), and thereby ensuring sufficient exposure to Old French during training. 

\paragraph{Experimental design}
We evaluate on our manually curated test set, as well as on the first 2,000 entries of the dataset introduced by \citet{schonhardt_bdd_abbreviations_2025}, for which the Schonhardt baseline model should be at an advantage. All experiments are evaluated using CER and word error rate (WER), using the Levenshtein edit distance normalized over the length of the ground truth.

\subsection{Results}
Evaluated on our gold dataset, our proposed model performs well for both languages, improving over the zero-shot model and the only other available model for the task by significant margins (Table~\ref{tab:cer}). On the Schonhardt dataset, our model shows a stronger performance than the original model itself, the original model showing a tendency to undergenerate. 

\begin{table}[]
    \centering
    \resizebox{\linewidth}{!}{
\begin{tabular}{lrrrrrr}
	\toprule
	Model      & \multicolumn{3}{c}{CER} & \multicolumn{3}{c}{WER} \\
	                 & All          & \textit{Latin} & \textit{Old French} & All           & \textit{Latin} & \textit{Old French} \\ \midrule
    \multicolumn{7}{c}{\textit{On our gold data}} \\
    \midrule
	Llama 3        & 139.9        & 160.4          & 79.9            & 163.7         & 183.1          & 107.3           \\
	ByT5 (Schonhardt)       & 64.8         & 75.6           & 33.3            & 92.3          & 104.8          & 55.9            \\
	ByT5 (Ours)   & \textbf{6.7} & \textbf{6.3}   & \textbf{7.6}             & \textbf{20.7} & \textbf{19.6}  & \textbf{23.8}            \\ \midrule
    \multicolumn{7}{c}{\textit{On Schonhardt data (Latin only)}} \\
    \midrule
    Llama 3 & & 26.4 &&&  54.5 & \\
    ByT5 (Schonhardt) &&  130.0 &&&  109.4 &\\
    ByT5 (Ours) &&  \textbf{11.1} &&& \textbf{38.5} &\\ \bottomrule
    \bottomrule
\end{tabular}}
    \caption{Comparison of PEN results.}
    \label{tab:cer}
\end{table}

Qualitative analysis reveals a recurrent \emph{classical bias} in the normalization of Latin, whereby the model tends to favor forms aligned with standardized spelling over historically attested medieval variants, except for final \character{-e} instead of \character{ae}. This tendency is illustrated by systematic preferences for \character{ph} over \character{f} and for \character{-tio-} over \character{-cio-}, reflecting a reversion to classical spellings. It also over-predicts names without context knowledge, such as \character{Johannis} for \character{Io.} and often struggles to adapt abbreviation resolution to the context, not unlike some of the models of Torres' work.

In Old French, the absence of stabilized orthographic conventions similarly leads the model to privilege dominant variants in its training set, normalizing forms such as \character{leurs} to \character{lors} and \character{pueple} to \character{pople}. This results in the systematic erasure of localized \emph{scripta} traditions, as the model prioritizes alignment targets over historically meaningful scribal variation. It however shows a strong capacity to retokenize or correct wrong ATR predictions.

Finally, we observe cases of variant insertion by the model, or at least semantic failure such as \character{iustum} for \character{iterum}. While this phenomenon constitutes a limitation of the baseline model introduced in this study, it nevertheless highlights the importance of decoupling PEN from the ATR process itself. From a data-centric perspective, this separation enables us to trace normalization errors back to pre-editorial stages without requiring direct recourse to the original manuscript sources.

\section{Impact on a Downstream Task}

Given these results and the observed overnormalization effects, a remaining question is whether our normalization model—trained on the alignment dataset—can nevertheless outperform a previously established pipeline for downstream text analysis. To address this question, we compare our approach with the pipeline proposed by \citet{wauchieristhatyou} in the context of stylometric authorship analysis of \textit{Vie de Saint Lambert}, focusing on the extraction of lemmas and PoS information, including PoS 3-grams.

\subsection{Setup}

Stylometric analysis of Old French texts is strongly hindered by spelling variation, particularly in inter-manuscript settings. Their pipeline addresses this issue by combining a convolutional neural network for space error correction with a model for both PoS-tagging and lemmatization \cite{clerice_thibault_2019_3237455} fine-tuned on artificially abbreviated data (hereafter, the \textit{Lambert} pipeline). On the other hand, our pipeline uses our normalization model and the same PoS-tagger/lemmatizer from \citep{clerice_thibault_2019_3237455}.

\paragraph{Data}
We conduct our evaluation on their transcription of \textit{Vie de Saint Lambert} in the manuscript BnF Fr. 412. We compare their lemmatization and PoS-tagging outputs and the annotations produced by our own pipeline against the ground truth annotation of the normalized version. The \textit{Vie} displays a very low amount of contextual abbreviation, and is mostly hindered by spacing issues and ATR errors. We compare the ground-truth based on the full results and on the 100 most frequent words (MFW), in a setting similar to stylometry practices (ignoring any punctuation).

\paragraph{Evaluation}
To assess the impact of these different preprocessing pipelines, we adopt a Bag-of-Words (BoW) evaluation framework and compute precision and recall in order to measure their suitability for frequency-based analyses. Given $G$ and $P$ the sequences of gold and predicted labels, and $c_G(x)$ and $c_P(x)$ the number of times a label $x$ occurs in $G$ and $P$, 
we define the number of true positives as
$\mathrm{TP} = \sum_{x} \min\bigl(c_G(x), c_P(x)\bigr)$. We then have $\mathrm{Precision}_{\mathrm{BoW}} = \frac{\mathrm{TP}}{|P|}$, and $\mathrm{Recall}_{\mathrm{BoW}} = \frac{\mathrm{TP}}{|G|}$.

\begin{table}[]
    \centering
    \footnotesize
    \resizebox{\linewidth}{!}{
\begin{tabular}{lrrrrrrrrr}
\toprule
	Pipeline      & \multicolumn{3}{c}{Lemma} & \multicolumn{3}{c}{POS} & \multicolumn{3}{c}{POS 3-grams}      \\ 
	Name    & Pre.          & Rec.          & F1            & Pre.          & Rec.           & F1            & Prec.         & Rec.          & F1            \\
\midrule
\multicolumn{10}{c}{\textit{All tokens}} \\
\midrule
Lambert &  88.2 &  \textbf{89.4} &  88.8 &  95.8 &  \textbf{97.2} &  96.5 &  85.6 &  \textbf{86.8} &  86.2 \\
Ours &  \textbf{90.9} &  89.0 &  \textbf{89.9} &  \textbf{98.0} &  95.9 &  \textbf{96.9} &  \textbf{87.7} &  85.8 &  \textbf{86.7} \\
\midrule
\multicolumn{10}{c}{\textit{100 most frequent words (MFW)}} \\
\midrule
Lambert &  92.0 &  \textbf{92.6} &  \textbf{92.3}  &   & &   &   && \\
Ours &  \textbf{94.3} &  89.5 &  91.8 &   & &   &   &&  \\
\bottomrule
\end{tabular}}
    \caption{Impact of the normalizer on downstream tasks (PoS tagging and lemmatization).}
    \label{tab:lemmatization}
\end{table}

\subsection{Results}

Overall, both models produce a comparable number of tokens. The Lambert pipeline generates approximately 2\% more tokens, whereas the ByT5-based approach results in 3.8\% fewer tokens than the edited text. For both systems, our error analysis reveals that annotation errors are not directly correlated with the correctness of the normalization step applied to individual tokens. For instance, the word \enquote{uie} (life), which is a non-ambiguous input for manual lemmatization, is poorly lemmatized 66\% of the time by the Lambert pipeline and 30\% by ours, with two out of three errors due to a non-normalization to \enquote{vie} and a lemmatization to \textsc{uie} instead of \textsc{vie1}.

The results (Table~\ref{tab:lemmatization}) exhibit a consistent general trend across all evaluated subsets. Precision and F1 are systematically higher for our approach, within a few percentage points, while recall is consistently higher for the Lambert pipeline by a comparable margin (except for MFW lemmatization). Our model introduces less noise than Lambert's pipeline but undergenerates and produces inadequate content for the tagger: for example, it does not produce apostrophes (\enquote{qu il} instead of \enquote{qu'il}), whereas they seem to be an overused marker by the tagger.

In the context of stylometric analysis, both approaches therefore present complementary advantages, depending on their respective impact on the distribution of the 100 most frequent words. The advantage of our approach lies in a potential simplification of the pipeline, contrary to the multiplication of steps and ad-hoc models from Lambert's.

\section{Discussion and Conclusion}

We introduced pre-editorial normalization (PEN) as an intermediate task between graphemic ATR output and fully edited (normalized) text, and presented a large-scale aligned dataset for medieval Latin and Old French. Our experiments show that models trained on this resource substantially outperform existing baselines on normalization, and that PEN shows promising results as a first step for downstream linguistic analysis while simplifying processing pipelines.

At the same time, our results highlight persistent risks of over-normalization and editorial bias, particularly in the tendency to favor more classical spellings for Latin and alternative ones for Old French. These findings underline the importance of treating normalization as an intermediate layer rather than an implicit part of transcription.

Future work will focus on expanding the dataset, modeling uncertainty in normalization, and more importantly page-wide and document-level normalization settings, in line with recent practices in machine translation to use broader contextual windows. We hope that this resource and task definition will support more transparent and reusable approaches to research on historical texts.


\nocite{*}
\section{Bibliographical References}\label{sec:reference}

\bibliographystyle{lrec2026-natbib}
\bibliography{bibliography}

\begin{thebibliography}{25}
\expandafter\ifx\csname natexlab\endcsname\relax\def\natexlab#1{#1}\fi

\bibitem[{Assmann and Sahle(1826--)}]{MGH}
Assmann, B. and Sahle, Patrick. 1826--.
\newblock \href {https://www.dmgh.de/} {\emph{Monumenta Germaniae Historica}}.

\bibitem[{Bauduin et~al.(2023)Bauduin, Buard, Goloubkoff, Mancel, Fournier-Fujimoto, Buquet, Brossard, Trohel, Combalbert, Bisson, Berthelot, Dubois, and Bloche}]{Scripta}
Bauduin, Pierre and Buard, Pierre-Yves and Goloubkoff, Anne and Mancel, Émeline and Fournier-Fujimoto, Tamiko and Buquet, Thierry and Brossard, Stéphane and Trohel, Benoît and Combalbert, Grégory and Bisson, Marie and Berthelot, Clémentine and Dubois, Adrien and Bloche, Michaël. 2023.
\newblock \href {https://mrsh.unicaen.fr/scripta/accueil.html} {\emph{Scripta}}.

\bibitem[{Camps et~al.(2019)Camps, Cochet, Ing, and Albarran}]{Geste}
Jean-Baptiste Camps and Alice Cochet and Lucence Ing and Elena Albarran. 2019.
\newblock \href {https://doi.org/10.5281/zenodo.2630574} {\emph{Geste: un corpus de chansons de geste, 2016-…}}.
\newblock Zenodo.

\bibitem[{Clérice(2021)}]{Clerice}
Clérice, Thibault. 2021.
\newblock \href {https://doi.org/10.5281/zenodo.4384732} {\emph{{Corpora of rare texts, Lasciva Roma}}}.

\bibitem[{Crane et~al.(2021{\natexlab{a}})Crane, Babeu, Cerrato, Munson, Dee, Gessner, Robertson, Franzini, Stoyanova, Selle, Springmann, and Clérice}]{csel}
Crane, Gregory Ralph and Babeu, Alison and Cerrato, Lisa and Munson, Matthew and Dee, Stella and Gessner, Annette and Robertson, Bruce and Franzini, Greta and Stoyanova, Simona and Selle, Tabea and Springmann, Uwe and Clérice, Thibault. 2021{\natexlab{a}}.
\newblock \href {https://github.com/OpenGreekAndLatin/csel-dev} {\emph{Corpus Scriptorum Ecclesiasticorum Latinorum digitized by Open Greek And Latin}}.

\bibitem[{Crane et~al.(2021{\natexlab{b}})Crane, Mylonas, Smith, Almas, Babeu, Berra, Bonhomme, Brooks, Cerrato, Cl\'erice, Dee, Ehwald, Frering, Gessner, Harrington, Himes, Jovanovic, Konieczny, Mahoney, Merrill, Mimno, Mueller, Munson, Singhal, Smith, Taounza-Jeminet, Weaver, and Wulfman}]{perseus}
Crane, Gregory Ralph and Mylonas, Elli and Smith, Neel and Almas, Bridget and Babeu, Alison and Berra, Aurélien and Bonhomme, Marie-Laurence and Brooks, David and Cerrato, Lisa and Cl\'erice, Thibault and Dee, Stella and Ehwald, R. and Frering, Léa and Gessner, Annette and Harrington, J. Matthew and Himes, Zach and Jovanovic, Neven and Konieczny, Michael and Mahoney, Anne and Merrill, William and Mimno, David and Mueller, Lucian and Munson, Matthew and Singhal, Rashmi and Smith, David and Taounza-Jeminet, Alyx and Weaver, Gabriel A. and Wulfman, C. 2021{\natexlab{b}}.
\newblock \href {https://github.com/PerseusDL/canonical-latinLit} {\emph{Perseus Canonical Latin Literature}}.

\bibitem[{Dees et~al.(2006)Dees, Stein, Kunstmann, Gleßgen, and Souvay}]{NCA}
Anthonij Dees and Achim Stein and Pierre Kunstmann and Martin Gleßgen and Giles Souvay. 2006.
\newblock \href {http://stella.atilf.fr/gsouvay/nca/} {\emph{Nouveau Corpus d'Amsterdam}}.

\bibitem[{Draelants and Kuhry(2020)}]{EncyMe}
Draelants, Isabelle and Kuhry, Emmanuelle. 2020.
\newblock \href {https://sourcencyme.irht.cnrs.fr/} {\emph{Sources des Encyclopédies Médiévale}}.

\bibitem[{Glessgen et~al.(2016)Glessgen, Carles, Duval, and Videsott}]{DocLing}
Glessgen, Martin-Dietrich and Carles, Hélène and Duval, Frédéric and Videsott, Paul. 2016.
\newblock \href {https://www.rose.uzh.ch/docling/} {\emph{Documents linguistiques galloromans}}.

\bibitem[{Goetz and Gitner()}]{CGL}
Goetz, Georg and Gitner, Adam, et al.
\newblock \emph{CGLO}.
\newblock 23-12.

\bibitem[{Guillot et~al.(2022)Guillot, Heiden, and Lavrentiev}]{BFM}
Guillot, C{\'e}line and Heiden, Serge and Lavrentiev, Alexei. 2022.
\newblock \href {https://nakala.fr/collection/10.34847/nkl.1279lie9} {\emph{Base de fran{\c{c}}ais m{\'e}di{\'e}val}}.

\bibitem[{Hasse(2023)}]{Ptolemaeus}
Hasse, Dag Nikolaus. 2023.
\newblock \href {https://ptolemaeus.badw.de/} {\emph{Ptolemaeus Arabus et Latinus}}.

\bibitem[{Jovanović(2024)}]{Croala}
Neven Jovanović. 2024.
\newblock \href {https://doi.org/10.5281/zenodo.10657786} {\emph{Croatiae Auctores Latini Textus}}.
\newblock Zenodo.

\bibitem[{Morcos et~al.(2020)Morcos, Gaunt, Rachetta, Ravenhall, Ventura, Barbieri, Caton, Noël, Ferraro, and Romanova}]{tvof}
Hannah Morcos and Simon Gaunt and Maria Teresa Rachetta and Henry Ravenhall and Simone Ventura and Luca Barbieri and Paul Caton and Geoffroy Noël and Ginestra Ferraro and Natasha Romanova. 2020.
\newblock \href {https://doi.org/10.6084/m9.figshare.11907081.v3} {\emph{{The Histoire ancienne jusqu’à César: A Digital Edition (The Values of French)}}}.

\bibitem[{Perreaux(2021)}]{perreauCEMA}
Perreaux, Nicolas. 2021.
\newblock \href {https://cema.lamop.fr} {\emph{Cartae Europae Medii Aevi (CEMA)}}.

\bibitem[{Pierreville et~al.(2025)Pierreville, Nuguet, Pinche, and Lavrentiev}]{fabliaux}
Corinne Pierreville and Jules Nuguet and Ariane Pinche and Alexei Lavrentiev. 2025.
\newblock \href {https://gitlab.huma-num.fr/fabliaux/} {\emph{Projet Fabliaux}}.

\bibitem[{Pinelli et~al.(2023)Pinelli, Ctibor, Roelli, Degl'Innocenti, Gamberini, Carlamaria~Crespi, Fizzarotti, Montepaone, Vangone, and Santarelli}]{Mirabile}
Pinelli, Luca and Ctibor, Jan and Roelli, Philipp and Degl'Innocenti, Antonella and Gamberini, Roberto and Carlamaria Crespi, Serena and Fizzarotti, Luisa and Montepaone, Olivia and Vangone, Laura and Santarelli, Riccardo. 2023.
\newblock \href {https://mdl.mirabileweb.it} {\emph{Mirabile Digital Archive}}.

\bibitem[{Roelli and Ctibor(2024)}]{CC}
Roelli, Philipp and Ctibor, Jan. 2024.
\newblock \href {https://mlat.uzh.ch/} {\emph{Corpus Corporum}}.

\bibitem[{Russo(2023)}]{Rinascimento}
Russo, Emilio. 2023.
\newblock \href {http://www.bibliotecaitaliana.it/} {\emph{Biblioteca Italiana}}.

\bibitem[{Schibel et~al.(1999--2008)Schibel, Kredel, and Niehl}]{NeoLatinitas}
Schibel, Wolfgang and Kredel, Heinz and Niehl, Rüdiger. 1999--2008.
\newblock \href {https://mateo.uni-mannheim.de/camenahtdocs/camena_e.html} {\emph{CAMENA: Latin Texts of Early Modern Europe}}.
\newblock DFG-funded project; online access.

\bibitem[{Sneddon(1982)}]{ofc}
Clive R. Sneddon. 1982.
\newblock \href {http://hdl.handle.net/20.500.14106/0176} {\emph{Old French corpus}}.
\newblock Literary and Linguistic Data Service.

\bibitem[{Stella et~al.(2025)Stella, D’Angelo, Maria, Buzzoni, Donne, and Turco}]{Latinitaitaliana}
Francesco Stella and Edoardo D’Angelo and Giorgio Di Maria and Marina Buzzoni and Fulvio Delle Donne and Roberto Rosselli Del Turco. 2025.
\newblock \href {https://en.alim.unisi.it/} {\emph{ALIM: Archivio della Latinità Italiana del Medioevo}}.
\newblock Accessed: 2025-09-27.

\bibitem[{Stutzmann et~al.(2021)Stutzmann, Torres~Aguilar, and Chaffenet}]{Alcar}
Stutzmann, Dominique and Torres Aguilar, Sergio and Chaffenet, Paul. 2021.
\newblock \href {https://doi.org/10.5281/zenodo.5600884} {\emph{HOME-Alcar: Aligned and Annotated Cartularies}}.
\newblock Zenodo.

\bibitem[{Tabacco et~al.(2021)Tabacco, Lana, Balossino, Bessi, Bognini, Buffa, Caso, Cattaneo, Ciotti, Ciusani, Colombo, Cuzzotti, del Core, Della~Calce, Denicola, Digirolamo, Ferrandi, Ferroni, Fontana, Guiglia, Loberti, Lucciano, Maconi, Malaspina, Manuela, Marini, Maronet, Massano, Mazzucco, Mellano, Miglietta, Mollea, Mosca, Musso, Naso, Paniagua, Poncina, Ramires, Rinaldi, Rosso, Rota, Rozzi, Rugnone, Senore, Stok, Strona, Verny, and Vittoria~Martino}]{digiliblt}
Tabacco, Raffaella and Lana, Maurizio and Balossino, Michele and Bessi, Giancarlo and Bognini, Filippo and Buffa, Martina and Caso, Daniela and Cattaneo, Gianmario and Ciotti, Fabio and Ciusani, Mauro and Colombo, Piero and Cuzzotti, Claudia and del Core, Vincenzo and Della Calce, Elisa and Denicola, Luciano and Digirolamo, Letizia and Ferrandi, Etienne and Ferroni, Manuela and Fontana, Davide and Guiglia, Valeria and Loberti , Martina and Lucciano, Melanie and Maconi, Ludovica and Malaspina, Ermanno and Manuela, Ferroni and Marini, Alessia and Maronet, Léa and Massano, Federico and Mazzucco, Clementina and Mellano, Anastasia and Miglietta, Chiara and Mollea, Simone and Mosca, Laura and Musso, Simona and Naso, Manuela and Paniagua, David and Poncina, Fabio and Ramires, Giuseppe and Rinaldi, Valentina and Rosso, Nadia and Rota, Simona and Rozzi, Stefano and Rugnone, Elisa and Senore, Corinna and Stok, Fabio and Strona, Beatrice and Verny, Romain and Vittoria Martino, Maria. 2021.
\newblock \href {https://digiliblt.uniupo.it/progetto.php} {\emph{Digital Library of Late-Antique Latin Texts}}.

\bibitem[{Théry(2022)}]{Aposcripta}
Théry, Julien. 2022.
\newblock \href {https://doi.org/10.5281/zenodo.6771270} {\emph{APOSCRIPTA database. Unified Corpus of Papal Letters}}.
\newblock Zenodo.

\end{thebibliography}

\section{Language Resource References}
\label{lr:ref}
\bibliographystylelanguageresource{lrec2026-natbib}
\bibliographylanguageresource{languageresource}

\finalonly{%
\appendix
\renewcommand{\thetable}{\Alph{section}.\arabic{table}}

\section{Dataset and models}

The ByT5 model can be found on HuggingFace at \url{https://huggingface.co/comma-project/normalization-byt5-small}. 

The dataset is available on HuggingFace at \url{https://huggingface.co/datasets/comma-project/pen-alignment-pairs}.

\section{Corpus Used for the Alignment}

\begin{table}[H]
    \centering
    \resizebox{\linewidth}{!}{\small
    \begin{tabular}{lr}
    \toprule
    Corpus (Latin then French) & \#kTokens \\
    \midrule
    Corpus Corporum (div.) \citeplanguageresource{CC} & 96,995 \\
    CEMA \citeplanguageresource{perreauCEMA} & 57,601 \\
    CAMENA \citeplanguageresource{NeoLatinitas} & 7,916 \\
    \citeplanguageresource{MGH} & 6,930 \\
    DigilibLT \citeplanguageresource{digiliblt} & 6,674 \\
    Perseus \citeplanguageresource{perseus} & 6,148 \\
    CSEL \citeplanguageresource{csel} & 6,046 \\
    CroaLa \citeplanguageresource{Croala} & 5,233 \\
    Mirabile \citeplanguageresource{Mirabile} & 4,655 \\
    Bib. Italiana \citeplanguageresource{Rinascimento} & 4,520 \\
    ALIM \citeplanguageresource{Latinitaitaliana} & 4,284 \\
    Add. Texts \citeplanguageresource{Clerice} & 3,844 \\
    Scripta \citeplanguageresource{Scripta} & 2,226 \\
    Aposcripta \citeplanguageresource{Aposcripta} & 1,736 \\
    ALCAR \citeplanguageresource{Alcar} & 973 \\
    Dig. Ptolemaeus \citeplanguageresource{Ptolemaeus} & 880 \\
    CGLO \citeplanguageresource{CGL} & 855 \\
    SourceEncyMe \citeplanguageresource{EncyMe} & 777 \\ \midrule
    NCA \citetlanguageresource{NCA} & 3,185 \\
    BFM \citeplanguageresource{BFM} & 6,293 \\
    Fabliaux \citeplanguageresource{fabliaux} & 120 \\
    Geste \citeplanguageresource{Geste} & 324 \\
    DocLing \citeplanguageresource{DocLing} & 1,278 \\
    TVOF \citeplanguageresource{tvof} & 361 \\
    OFC \citeplanguageresource{ofc} & 14 \\
    \midrule
    Total & 229,870 \\
    \bottomrule
    \end{tabular}}
    \caption{Normalized corpus used by CoMMA for the ModernBert model.}
    \label{tab:sources}
\end{table}

\section{Ranking of Corpora according to the Number of Alignments}

\begin{table}[H]
\centering
\small
\begin{tabular}{lr}
\toprule
Corpus                         & \#alignments \\ \midrule
Patrologia Latina              & 2,838,542    \\
Lasciva Roma                   & 413,580     \\
CSEL                           & 377,601     \\
PerseusDL                      & 306,297     \\
DigilibLT                      & 179,899     \\
Auctores Scientiarum Varii     & 110,535     \\
CEMA-latin                     & 78,570      \\ \bottomrule
\end{tabular}
\caption{The most frequent aligned corpora in the dataset.}
\label{tab:corpus-aligned}
\end{table}

\section{Llama3 8B Prompt}
\label{app:prompt}

\textbf{system:} You are a specialist in Medieval Paleography and HTR correction. 

Your goal is to transform "raw" HTR transcriptions into normalized Medieval Old French or Latin.

\#\#\# LINGUISTIC \& PALAEOGRAPHIC RULES:

\begin{itemize}
    \item HTR ERROR CORRECTION: Fix character confusions typical of HTR, such as `m' being misread for `in' (e.g., `mexper' -> `inexper') or `f' for `s'.
    \item U/V \& I/J NORMALIZATION: Distinguish between vocalic and consonantal uses. Use `v' and `j' for consonants and `u' and `i' for vowels.
    \item EXPANSION: Fully expand all scribal abbreviations and suspensions into their full character forms.
    \item GRAPHICAL VARIATION: Fully respect graphical variation, such as filosofia should be kept instead of normalizing to philosophia.
    \item APOSTROPHE: Add apostrophe where necessary (e.g., `dun' -> `d'un').
    \item SPACES: Remove extra spaces (e.g., `co eur' -> `coeur') and add necessary ones (e.g., `letriste' -> `le triste').
\end{itemize}

\#\#\# ABBREVIATION KEY:

\begin{itemize}
    \item {\junicodeFont 9 or ꝯ} = com / con (e.g., {\junicodeFont 9}me -> comme)
    \item {\junicodeFont ꝰ} = us (e.g., vo{\junicodeFont ꝰ} -> vous)
    \item {\junicodeFont qͥ} = qui
    \item ł = ol (e.g., młt -> molt)
    \item {\junicodeFont ◌̃} (tilde/macron) = nasalization (n/m) or missing letters (e.g., q̃  -> que)
    \item vertical tilde for suspension (missing letters) or R based phonemes (e.g., -re/er/-ra/ar-)
\end{itemize}

    Return ONLY JSON.
    
    \textbf{user:}    Normalize this HTR output: {{ text }}
    
    \textbf{assistant:}
}
\end{document}